\documentclass[11pt,twocolumn]{article}

\usepackage[a4paper,margin=2.5cm]{geometry}
\usepackage[T1]{fontenc}
\usepackage{times}
\usepackage{microtype}

\newcommand{\arb}[1]{\textarab{#1}}
\newcommand{\textarab}[1]{{\fontfamily{cmr}\selectfont[#1]}}

\usepackage{amsmath}
\usepackage{amssymb}
\usepackage{graphicx}
\usepackage{booktabs}
\usepackage{multirow}
\usepackage{xcolor}
\usepackage{hyperref}
\usepackage{url}
\usepackage{subcaption}
\usepackage[font=small,labelfont=bf]{caption}

\usepackage{titlesec}
\titleformat{\section}{\large\bfseries}{\thesection}{1em}{}
\titleformat{\subsection}{\normalsize\bfseries}{\thesubsection}{1em}{}
\titlespacing*{\section}{0pt}{2.5ex plus 1ex minus .2ex}{1.5ex plus .2ex}
\titlespacing*{\subsection}{0pt}{2ex plus 0.8ex minus .2ex}{1ex plus .2ex}

\hypersetup{
    colorlinks=true,
    linkcolor=blue,
    filecolor=magenta,
    urlcolor=cyan,
    citecolor=blue,
    pdftitle={AraToken: Arabic Tokenizer Optimization with Language Extension Pipeline},
    pdfauthor={Mark Kashirskiy, Artiom Lipinski, Ilya Makarov},
}

\title{AraToken: Optimizing Arabic Tokenization with Normalization Pipeline and Language Extension for Qwen3}

\author{
    Mark Kashirskiy$^{1,2}$ \quad Artiom Lipinski$^{3}$ \quad Ilya Makarov$^{4}$ \\[1ex]
    $^{1}$Higher School of Economics, Moscow, Russia, Moscow, Russia \\
    $^{2}$AI Talent Hub, ITMO University, Saint Petersburg, Russia \\
    $^{3}$Markov Lab, Saint Petersburg State University, Russia \\
    $^{4}$Higher School of Economics, Moscow, Russia \\[0.5ex]
    \texttt{353056@niuitmo.ru} \quad \texttt{st107742@student.spbu.ru} \quad \texttt{iamakarov@hse.ru}
}

\date{}

\begin{document}

\maketitle

\begin{abstract}
Tokenization is a critical preprocessing step for large language models (LLMs), directly impacting training efficiency and downstream performance. General-purpose tokenizers trained predominantly on English and Latin-script languages exhibit suboptimal performance on morphologically rich languages such as Arabic, resulting in inflated token sequences and reduced compression efficiency. In this work, we present AraToken, an Arabic-optimized tokenizer built on SentencePiece Unigram algorithm with a comprehensive normalization pipeline addressing Arabic-specific orthographic variations including Alif variants, diacritics, and Arabic-Indic numerals. We systematically compare BPE, WordPiece, and SentencePiece algorithms across multiple configurations, demonstrating that SentencePiece with normalization achieves 18\% lower fertility (1.199 vs 1.35 tokens/word) compared to unnormalized baselines. Furthermore, we introduce the Language Extension Pipeline (LEP), a method for integrating the optimized tokenizer into Qwen3-0.6B through vocabulary extension with mean subtoken initialization and selective transformer layer unfreezing. Our experiments show that LEP reduces evaluation loss from 8.28 to 2.43 within 800 training steps on 100K Arabic samples. We release our tokenizer, training scripts, and model checkpoints to facilitate Arabic NLP research.
\end{abstract}

\section{Introduction}
\label{sec:introduction}

Large language models (LLMs) have demonstrated remarkable capabilities across a wide range of natural language processing tasks \cite{brown2020language, touvron2023llama, qwen2024qwen2}. However, the effectiveness of these models is fundamentally constrained by their tokenization strategy. Tokenizers trained on predominantly English corpora often exhibit poor compression efficiency for non-Latin scripts and morphologically rich languages \cite{rust2021good, petrov2023language}.

Arabic presents unique challenges for tokenization due to several linguistic characteristics. First, Arabic is a highly inflected language where words carry extensive morphological information through prefixes, suffixes, and infixes \cite{habash2010introduction}. Second, Arabic orthography exhibits significant variability, particularly in the representation of Alif variants (\arb{Hamza-above}, \arb{Hamza-below}, \arb{Madda}, \arb{Alif}) and the optional nature of diacritical marks (harakat). Third, Arabic text frequently contains Arabic-Indic numerals and specialized punctuation that require explicit normalization.

These challenges result in general-purpose tokenizers producing excessively fragmented token sequences for Arabic text, leading to: (1) increased computational costs during training and inference, (2) reduced effective context length, and (3) potential degradation in model performance on Arabic tasks.

In this paper, we address these challenges through a two-pronged approach:

\begin{itemize}
    \item \textbf{Arabic-Optimized Tokenizer}: We develop AraToken, a SentencePiece Unigram tokenizer trained on Arabic corpora with a comprehensive normalization pipeline that unifies orthographic variations and removes optional diacritics.
    
    \item \textbf{Language Extension Pipeline (LEP)}: We propose a method for integrating the optimized tokenizer into existing LLMs (specifically Qwen3) through vocabulary extension, mean subtoken initialization, and selective layer unfreezing.
\end{itemize}

Our experiments demonstrate that the normalized SentencePiece tokenizer achieves a fertility of 1.199 tokens per word, representing an 18\% improvement over unnormalized baselines. When integrated into Qwen3-0.6B via LEP, the model achieves an evaluation loss of 2.43 after only 800 training steps, compared to 8.28 without adaptation.

\begin{figure}[t]
\centering
\includegraphics[width=\columnwidth]{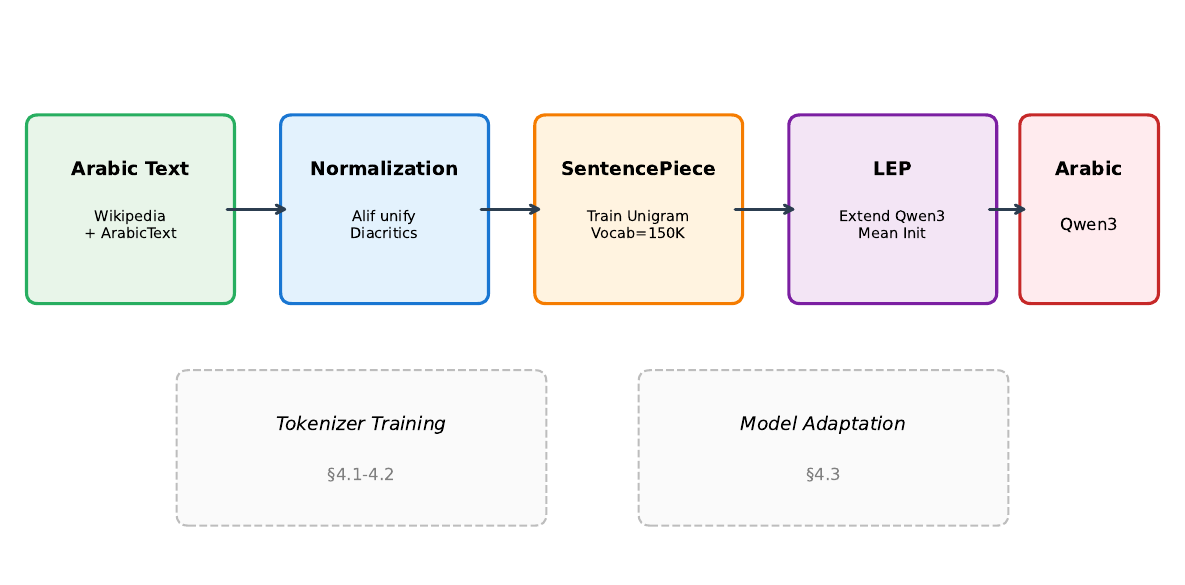}
\caption{Overview of the AraToken pipeline. Arabic text is normalized and used to train a SentencePiece tokenizer, which is then integrated into Qwen3 via the Language Extension Pipeline (LEP).}
\label{fig:overview}
\end{figure}

Figure~\ref{fig:overview} illustrates our overall approach, combining tokenizer training with model adaptation through LEP.

The remainder of this paper is organized as follows: Section~\ref{sec:related} reviews related work on tokenization and language adaptation. Section~\ref{sec:methodology} describes our normalization pipeline, tokenizer training, and LEP architecture. Section~\ref{sec:experiments} presents our experimental setup, and Section~\ref{sec:results} discusses the results. We conclude in Section~\ref{sec:conclusion} with limitations and future directions.

\section{Related Work}
\label{sec:related}

\subsection{Subword Tokenization Algorithms}

Modern LLMs predominantly employ subword tokenization to balance vocabulary size with coverage. Byte Pair Encoding (BPE) \cite{sennrich2016neural} iteratively merges the most frequent character pairs to construct a vocabulary. WordPiece \cite{schuster2012japanese} uses a likelihood-based criterion for merge decisions, while the Unigram algorithm \cite{kudo2018subword} learns a probabilistic language model over subword sequences using the EM algorithm.

SentencePiece \cite{kudo2018sentencepiece} provides a language-agnostic implementation supporting both BPE and Unigram algorithms, operating directly on raw text without pre-tokenization. This is particularly advantageous for languages like Arabic that do not use whitespace consistently.

\subsection{Arabic Natural Language Processing}

Arabic NLP has received significant attention due to the language's morphological complexity and dialectal variation \cite{darwish2021panoramic}. CAMeL Tools \cite{obeid2020camel} provides comprehensive utilities for Arabic preprocessing including morphological analysis and normalization. AraBART \cite{kamal2021arabart} and AraT5 \cite{nagoudi2022arat5} are pretrained transformer models specifically designed for Arabic, employing custom tokenization strategies.

Normalization is a critical preprocessing step for Arabic text \cite{habash2010introduction}. Common normalization operations include Alif unification (collapsing \arb{Hamza-above}, \arb{Hamza-below}, \arb{Madda} to \arb{Alif}), Hamza normalization, Ta Marbuta/Ha unification, and diacritic removal. The optimal normalization strategy depends on the downstream task, with some applications benefiting from preserved orthographic distinctions.

\subsection{Vocabulary Extension and Language Adaptation}

Extending pretrained LLMs to new languages has been explored through several approaches. BLOOM+1 \cite{yong2023bloom} investigates language adaptation strategies including continued pretraining and adapter-based methods, finding that adapters outperform continued pretraining for larger models. LLaMA Beyond English \cite{zhao2024llama} studies vocabulary extension for Chinese, demonstrating that effective transfer can be achieved with less than 1\% of the original pretraining data.

WECHSEL \cite{minixhofer2022wechsel} proposes cross-lingual embedding initialization for vocabulary extension, while FOCUS \cite{dobler2023focus} introduces a method for initializing new token embeddings based on semantic similarity. Our work builds on these approaches by combining vocabulary extension with selective layer unfreezing for Arabic adaptation.

\section{Methodology}
\label{sec:methodology}

\subsection{Arabic Normalization Pipeline}
\label{sec:normalization}

We implement a comprehensive Arabic normalization pipeline designed to reduce orthographic variability while preserving semantic content. The pipeline is integrated into the tokenizer's preprocessing stage using the HuggingFace Tokenizers library.

\paragraph{Unicode Normalization} We apply NFKC normalization as the first step to decompose compatibility characters and ensure consistent Unicode representation.

\paragraph{Alif Variant Unification} Arabic exhibits four common Alif variants that are often used interchangeably:
\begin{itemize}
    \item \arb{Hamza-above} (Hamza above) $\rightarrow$ \arb{Alif} (bare Alif)
    \item \arb{Hamza-below} (Hamza below) $\rightarrow$ \arb{Alif} (bare Alif)
    \item \arb{Madda} (Madda) $\rightarrow$ \arb{Alif} (bare Alif)
    \item Alif Wasla (U+0671) $\rightarrow$ \arb{Alif} (bare Alif)
\end{itemize}

We also experiment with preserving Alif variants (Alif4 configuration) to evaluate the trade-off between normalization and linguistic fidelity.

\paragraph{Numeral and Punctuation Normalization} Arabic-Indic numerals (\arb{0-9 Arabic}) are mapped to their Western Arabic equivalents (0-9). Arabic-specific punctuation marks (\arb{?-ar}, \arb{;-ar}, \arb{,-ar}) are normalized to their Latin counterparts.

\paragraph{Tatweel Removal} The Tatweel character (\arb{tatweel}) used for text justification is removed entirely.

\paragraph{Diacritics Handling} We provide two configurations: (1) drop diacritics, which removes all harakat for maximum normalization, and (2) keep diacritics, which preserves vowel marks for applications requiring phonetic information.

\begin{figure}[t]
\centering
\includegraphics[width=\columnwidth]{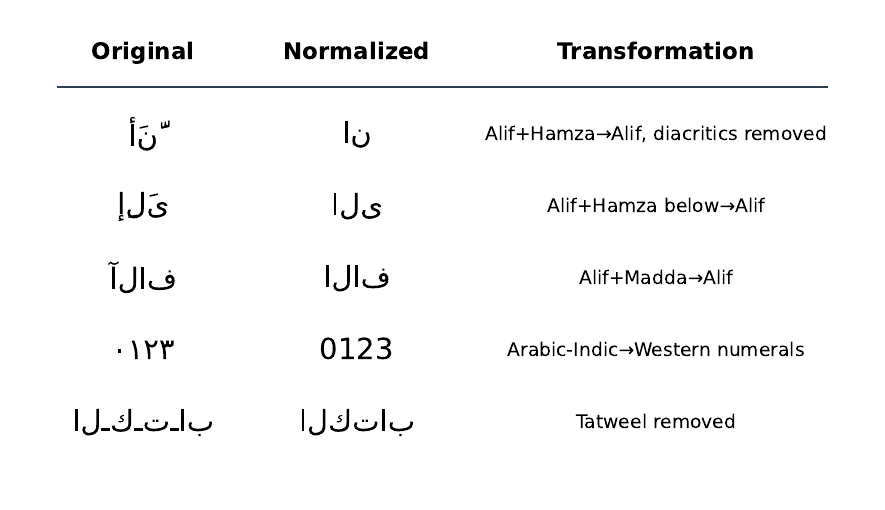}
\caption{Examples of Arabic text normalization. The pipeline unifies Alif variants, converts Arabic-Indic numerals to Western digits, and removes Tatweel (text justification) characters.}
\label{fig:normalization}
\end{figure}

Table~\ref{tab:normalization} summarizes the character replacement rules implemented in our normalization pipeline. Figure~\ref{fig:normalization} shows examples of text before and after normalization.

\begin{table}[t]
\centering
\small
\begin{tabular}{lcc}
\toprule
\textbf{Category} & \textbf{Original} & \textbf{Normalized} \\
\midrule
\multirow{4}{*}{Alif Variants} & \arb{Hamza-above} & \arb{Alif} \\
 & \arb{Hamza-below} & \arb{Alif} \\
 & \arb{Madda} & \arb{Alif} \\
 & Wasla & \arb{Alif} \\
\midrule
\multirow{3}{*}{Numerals} & \arb{0-9 Arabic} & 0--9 \\
 & \arb{dec-sep} & . \\
 & \arb{thou-sep} & , \\
\midrule
\multirow{3}{*}{Punctuation} & \arb{?-ar} & ? \\
 & \arb{;-ar} & ; \\
 & \arb{,-ar} & , \\
\midrule
Tatweel & \arb{tatweel} & (removed) \\
\bottomrule
\end{tabular}
\caption{Arabic character normalization rules. The pipeline applies these replacements sequentially after NFKC Unicode normalization.}
\label{tab:normalization}
\end{table}

\subsection{Tokenizer Training}
\label{sec:tokenizer_training}

We train tokenizers using three algorithms: BPE, WordPiece, and SentencePiece Unigram. For each algorithm, we explore configurations with and without normalization, and with dropped or retained diacritics.

\paragraph{Training Corpus} We construct our training corpus from two sources: (1) Arabic Wikipedia (approximately 1.1M articles), and (2) ArabicText-Large, a curated collection of Arabic web text. The combined corpus contains approximately 120 million tokens.

\paragraph{Vocabulary Size} We train tokenizers with a target vocabulary size of 80,000 tokens for base experiments and 150,000 tokens for normalized variants, matching the vocabulary scale of Qwen3.

\paragraph{Vocabulary Pruning} Following training, we apply frequency-based pruning to remove tokens covering less than 0.01\% of the corpus. We experiment with retention thresholds of 95\% and 99\% cumulative frequency coverage, resulting in pruned vocabularies of approximately 42K and 76K tokens respectively.

\paragraph{Evaluation Metrics} We evaluate tokenizers using three intrinsic metrics:
\begin{itemize}
    \item \textbf{Fertility}: Average number of tokens produced per word, where lower values indicate more efficient encoding.
    \item \textbf{Compression Ratio}: Ratio of characters to tokens, where higher values indicate better compression.
    \item \textbf{OOV Rate}: Percentage of words containing unknown tokens after tokenization.
\end{itemize}

\subsection{Language Extension Pipeline (LEP)}
\label{sec:lep}

\begin{figure}[t]
\centering
\includegraphics[width=\columnwidth]{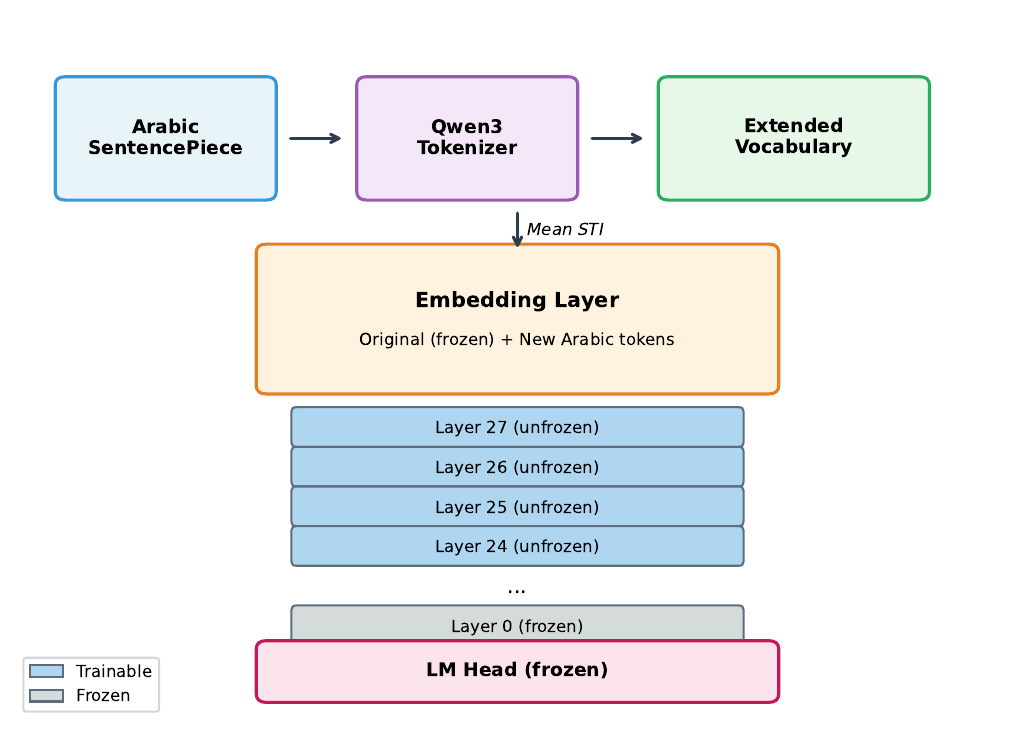}
\caption{Language Extension Pipeline (LEP) architecture. Arabic tokens from the SentencePiece model are added to the Qwen3 vocabulary. New embeddings are initialized using mean subtoken initialization. During training, original embeddings are frozen via gradient masking, while the last 4 transformer layers are unfrozen for adaptation.}
\label{fig:lep_architecture}
\end{figure}

The Language Extension Pipeline (LEP) integrates the Arabic-optimized tokenizer into Qwen3 through vocabulary extension and targeted fine-tuning. Figure~\ref{fig:lep_architecture} illustrates the overall architecture.

\paragraph{Vocabulary Extension} Given the trained SentencePiece model $\mathcal{V}_{ar}$ and the Qwen3 tokenizer $\mathcal{V}_{qwen}$, we extract Arabic tokens from $\mathcal{V}_{ar}$ that are not present in $\mathcal{V}_{qwen}$. Tokens beginning with the SentencePiece word boundary marker are added with the \texttt{lstrip=True} flag to handle whitespace correctly.

We filter tokens matching undesirable patterns (Latin characters, digits, Cyrillic, common punctuation) to ensure only Arabic-specific tokens are added.

\paragraph{Mean Subtoken Initialization} New token embeddings are initialized using the mean of their constituent subtoken embeddings from the original tokenizer:
\begin{equation}
\mathbf{e}_{new} = \frac{1}{|S|} \sum_{i \in S} \mathbf{e}_i
\end{equation}
where $S$ is the set of token IDs produced by encoding the new token string with the original Qwen3 tokenizer, and $\mathbf{e}_i$ are the corresponding embeddings.

This initialization provides a semantically meaningful starting point, as the new token's embedding is positioned near the centroid of its subtokens in embedding space.

\paragraph{Gradient Masking} During training, we freeze the embeddings of original Qwen3 tokens to prevent catastrophic forgetting. This is implemented through gradient hooks that zero out gradients for token indices below the vocabulary extension threshold:
\begin{equation}
\nabla_{\mathbf{e}_i} \mathcal{L} = 
\begin{cases}
0 & \text{if } i < |\mathcal{V}_{qwen}| \\
\nabla_{\mathbf{e}_i} \mathcal{L} & \text{otherwise}
\end{cases}
\end{equation}

\paragraph{Selective Layer Unfreezing} While the majority of transformer layers remain frozen, we unfreeze the last $k$ layers to allow the model to adapt its representations to the new tokenization. In our experiments, we unfreeze layers 24-27 (the final 4 layers of Qwen3-0.6B's 28-layer architecture).

\section{Experimental Setup}
\label{sec:experiments}

\subsection{Datasets}

\paragraph{Tokenizer Training} We use Arabic Wikipedia and ArabicText-Large for tokenizer training. The combined corpus contains approximately 120 million tokens after preprocessing.

\paragraph{LEP Training} For LEP experiments, we use a subset of ArabicText-Large containing 100,000 training samples and 2,000 validation samples. The text is preprocessed using our normalization pipeline before tokenization.

\subsection{Base Model}

We use Qwen3-0.6B-Base as our foundation model. Qwen3 employs byte-level BPE tokenization with a vocabulary size of 151,646 tokens and supports a context length of 32,768 tokens. The model architecture consists of 28 transformer layers with hidden dimension 1,024.

\subsection{Training Configuration}

Table~\ref{tab:training_config} summarizes our LEP training configuration. We use a linear learning rate schedule with 10\% warmup ratio and AdamW optimizer with weight decay 0.01. Gradient checkpointing is enabled to reduce memory consumption.

\begin{table}[t]
\centering
\small
\begin{tabular}{ll}
\toprule
\textbf{Parameter} & \textbf{Value} \\
\midrule
Base model & Qwen3-0.6B-Base \\
Learning rate & 2e-4 \\
LR scheduler & Linear \\
Warmup ratio & 0.1 \\
Batch size & 16 \\
Gradient accumulation & 6 \\
Max steps & 800 \\
Sequence length & 256 \\
Weight decay & 0.01 \\
Unfrozen layers & 24, 25, 26, 27 \\
Freeze old embeddings & True \\
\bottomrule
\end{tabular}
\caption{LEP training configuration.}
\label{tab:training_config}
\end{table}

\subsection{Evaluation Metrics}

For tokenizer evaluation, we compute fertility, compression ratio, and OOV rate on held-out Arabic text. For LEP training, we track training loss and evaluation loss on the validation set. Evaluation is performed every 125 steps.

\section{Results}
\label{sec:results}

\subsection{Tokenizer Comparison}

Table~\ref{tab:tokenizer_results} presents the intrinsic evaluation metrics for all tokenizer configurations. We observe several key findings:

\paragraph{SentencePiece outperforms BPE and WordPiece} Across all configurations, SentencePiece Unigram achieves the lowest fertility and highest compression ratio. With normalization, SentencePiece achieves fertility of 1.199 compared to 1.243 for BPE and 1.244 for WordPiece.

\paragraph{Normalization significantly improves efficiency} Applying our normalization pipeline reduces fertility by 8-9\% across all algorithms. For SentencePiece, normalization reduces fertility from 1.311 to 1.199 (8.5\% improvement).

\paragraph{Dropping diacritics further improves compression} Configurations that remove diacritics achieve better compression than those retaining them, as the reduced character set allows for more efficient subword segmentation.

\paragraph{OOV rates are negligible} All byte-level tokenizers (BPE, WordPiece) achieve 0\% OOV rate. SentencePiece shows a small OOV rate of approximately 0.1\%, which is acceptable for practical applications.

\begin{figure}[t]
\centering
\includegraphics[width=\columnwidth]{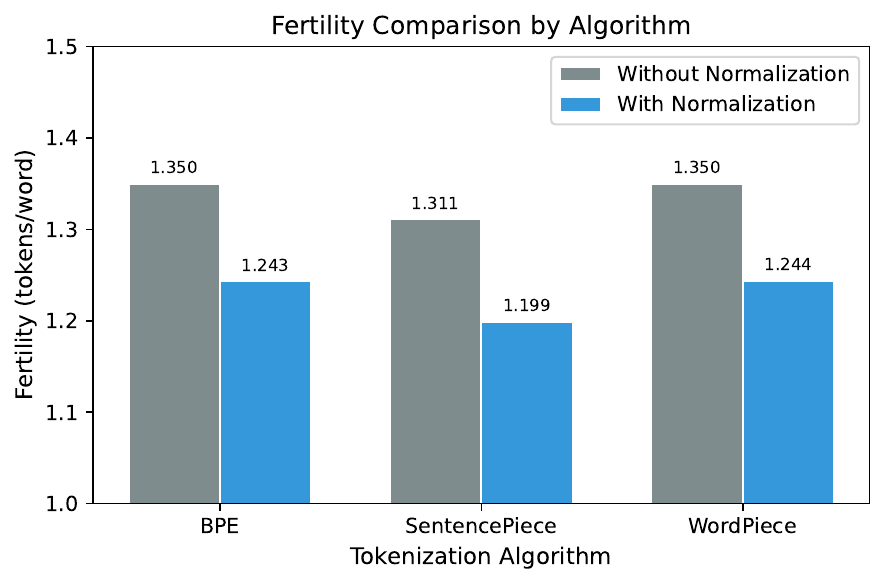}
\caption{Fertility comparison across tokenization algorithms. Lower fertility indicates more efficient encoding. Normalization consistently reduces fertility by 8-9\% across all algorithms.}
\label{fig:fertility}
\end{figure}

\begin{figure}[t]
\centering
\includegraphics[width=\columnwidth]{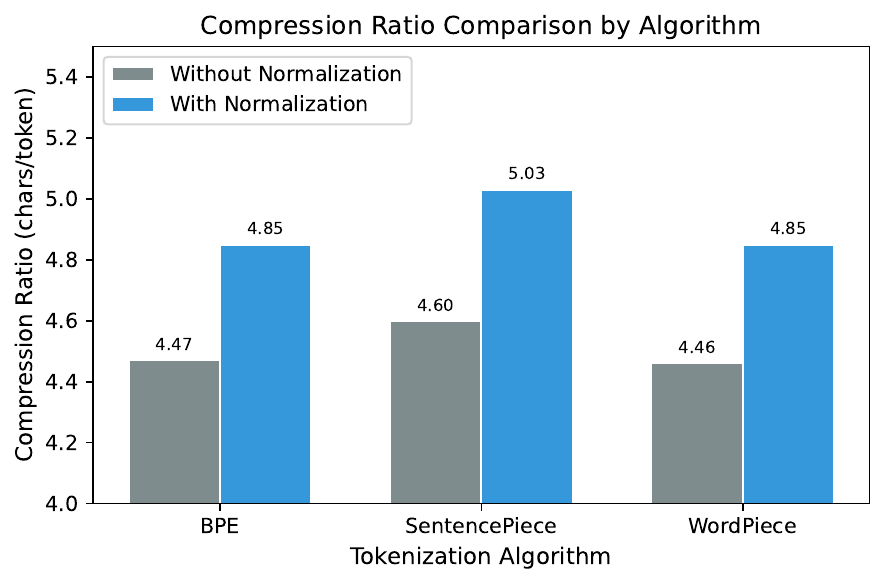}
\caption{Compression ratio comparison across tokenization algorithms. Higher compression indicates fewer tokens per character. SentencePiece with normalization achieves the best compression (5.03 chars/token).}
\label{fig:compression}
\end{figure}

Figures~\ref{fig:fertility} and~\ref{fig:compression} visualize the algorithm comparison with and without normalization.

\begin{table}[t]
\centering
\small
\begin{tabular}{lccc}
\toprule
\textbf{Tokenizer} & \textbf{Fertility} & \textbf{Comp.} & \textbf{OOV\%} \\
\midrule
\multicolumn{4}{l}{\textit{Drop Diacritics}} \\
bpe\_drop & 1.350 & 4.47 & 0.00 \\
bpe\_drop\_norm & 1.243 & 4.85 & 0.00 \\
sp\_drop & 1.311 & 4.60 & 0.10 \\
sp\_drop\_norm & \textbf{1.199} & \textbf{5.03} & 0.10 \\
wp\_drop & 1.350 & 4.46 & 0.00 \\
wp\_drop\_norm & 1.244 & 4.85 & 0.00 \\
\midrule
\multicolumn{4}{l}{\textit{Keep Diacritics}} \\
bpe\_keep & 1.415 & 4.29 & 0.00 \\
bpe\_keep\_norm & 1.308 & 4.64 & 0.00 \\
sp\_keep & 1.340 & 4.53 & 0.10 \\
sp\_keep\_norm & 1.218 & 4.98 & 0.10 \\
wp\_keep & 1.417 & 4.28 & 0.00 \\
wp\_keep\_norm & 1.309 & 4.64 & 0.00 \\
\bottomrule
\end{tabular}
\caption{Tokenizer evaluation results. Fertility is tokens per word (lower is better), Compression is characters per token (higher is better). ``norm'' indicates normalization applied.}
\label{tab:tokenizer_results}
\end{table}

\subsection{Vocabulary Pruning}

\begin{figure}[t]
\centering
\includegraphics[width=\columnwidth]{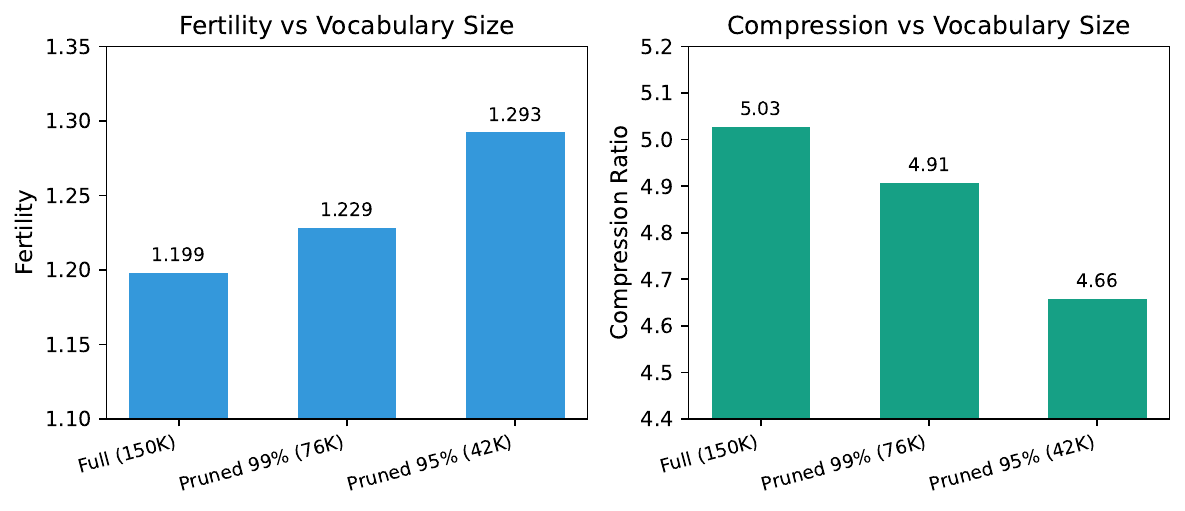}
\caption{Effect of vocabulary pruning on fertility and compression ratio. The 99\% coverage (76K) provides a good trade-off between vocabulary size and tokenization quality.}
\label{fig:pruning}
\end{figure}

Table~\ref{tab:pruning_results} and Figure~\ref{fig:pruning} show the effect of vocabulary pruning on tokenizer metrics. Pruning to 99\% coverage (76K vocabulary) maintains comparable performance to the full 150K vocabulary, while pruning to 95\% coverage (42K vocabulary) incurs a modest increase in fertility.

\begin{table}[t]
\centering
\small
\begin{tabular}{lccc}
\toprule
\textbf{Configuration} & \textbf{Vocab} & \textbf{Fertility} & \textbf{Comp.} \\
\midrule
sp\_drop\_norm & 150K & 1.199 & 5.03 \\
sp\_drop\_norm\_0.99 & 76K & 1.229 & 4.91 \\
sp\_drop\_norm\_0.95 & 42K & 1.293 & 4.66 \\
\bottomrule
\end{tabular}
\caption{Effect of vocabulary pruning on tokenizer metrics.}
\label{tab:pruning_results}
\end{table}

\subsection{LEP Training Results}

\begin{figure}[t]
\centering
\includegraphics[width=\columnwidth]{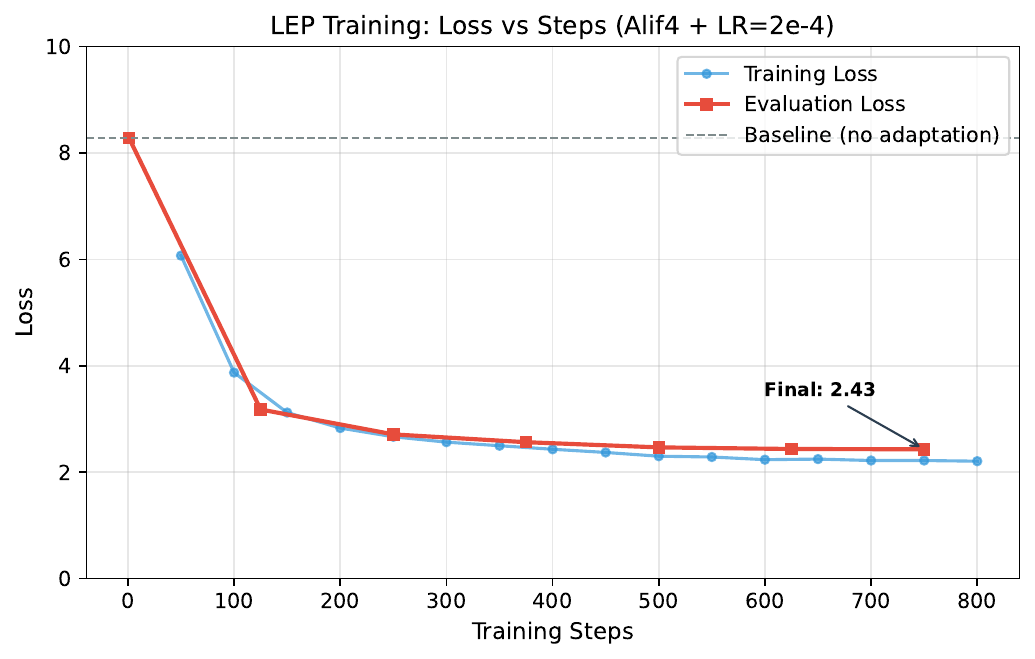}
\caption{LEP training curves. Evaluation loss decreases from 8.28 to 2.43 over 800 steps, demonstrating rapid adaptation to Arabic text.}
\label{fig:training_curves}
\end{figure}

Figure~\ref{fig:training_curves} shows the training and evaluation loss curves for LEP training. The model rapidly adapts to Arabic text, with evaluation loss decreasing from 8.28 to 2.43 within 800 training steps.

\paragraph{Ablation Studies} Table~\ref{tab:ablation} presents ablation results comparing different LEP configurations:

\begin{itemize}
    \item \textbf{Alif4 vs Unified}: Preserving Alif variants (Alif4) achieves lower final loss (2.43) compared to unified normalization (3.03), suggesting that orthographic distinctions carry useful information for language modeling.
    
    \item \textbf{Learning Rate}: Higher learning rate (2e-4) outperforms lower rate (8e-5), enabling faster adaptation within the limited training budget.
    
    \item \textbf{Layer Unfreezing}: Unfreezing the last 4 layers is essential for adaptation; freezing all transformer layers results in significantly higher loss.
\end{itemize}

\begin{table}[t]
\centering
\small
\begin{tabular}{lcc}
\toprule
\textbf{Configuration} & \textbf{Eval Loss} & \textbf{$\Delta$} \\
\midrule
Baseline (no adaptation) & 8.28 & -- \\
\midrule
Unified Alif, LR=2e-4 & 3.03 & -63\% \\
Alif4, LR=8e-5 & 4.78 & -42\% \\
Alif4, LR=2e-4 & \textbf{2.43} & \textbf{-71\%} \\
Alif4, LR=2e-4, freeze all & 4.02 & -51\% \\
\bottomrule
\end{tabular}
\caption{LEP ablation study. Alif4 with high learning rate and selective layer unfreezing achieves the best results.}
\label{tab:ablation}
\end{table}

\begin{figure}[t]
\centering
\includegraphics[width=\columnwidth]{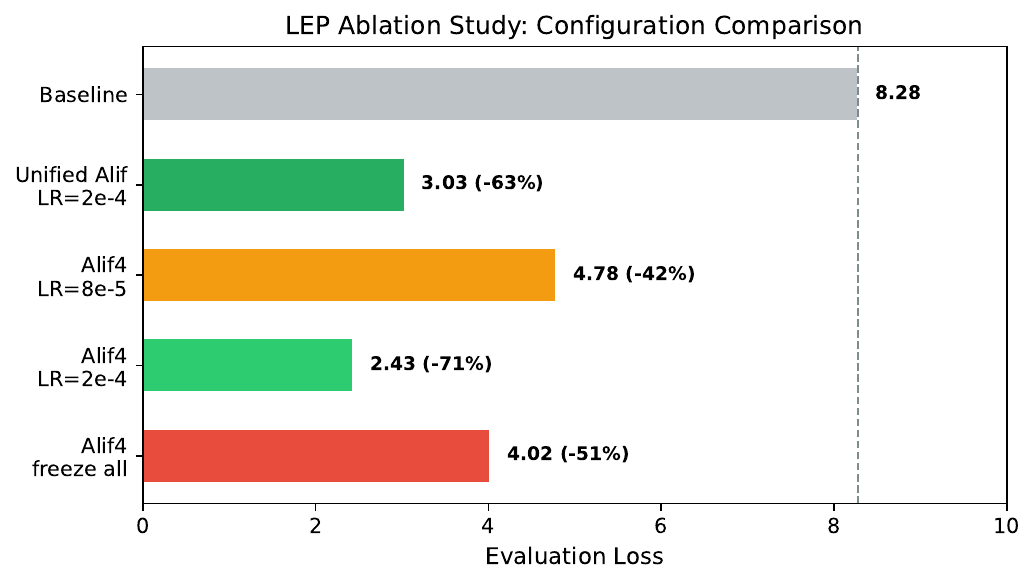}
\caption{Ablation study visualization showing evaluation loss for different LEP configurations. The best configuration (Alif4 + LR=2e-4) achieves 71\% reduction from baseline.}
\label{fig:ablation}
\end{figure}

Figure~\ref{fig:ablation} provides a visual comparison of ablation configurations.

\section{Analysis and Discussion}
\label{sec:discussion}

\subsection{Why SentencePiece Outperforms BPE/WordPiece}

SentencePiece's Unigram algorithm achieves superior compression for Arabic due to its probabilistic approach to segmentation. Unlike BPE's greedy merge strategy, Unigram considers the likelihood of the entire token sequence, enabling more globally optimal segmentations. This is particularly beneficial for Arabic's rich morphology, where multiple valid segmentations exist for inflected forms.

\subsection{Alif Variant Preservation}

Our experiments reveal a surprising finding: preserving Alif variants (Alif4 configuration) leads to lower language modeling loss compared to aggressive normalization. This suggests that Alif variants carry disambiguating information that aids language modeling. For example, the Hamza placement (\arb{Hamza-above} vs \arb{Hamza-below}) often indicates grammatical case or word origin.

This finding aligns with recent work questioning aggressive text normalization for neural models \cite{schwartz2019right}. We recommend that practitioners carefully consider the trade-off between tokenization efficiency and linguistic fidelity based on their downstream tasks.

\subsection{Efficiency of LEP}

The Language Extension Pipeline demonstrates remarkable efficiency, achieving significant adaptation within only 800 training steps on 100K samples. This represents less than 0.01\% of a typical LLM pretraining budget. Key factors contributing to this efficiency include:

\begin{enumerate}
    \item \textbf{Mean subtoken initialization}: Provides semantically meaningful starting points for new embeddings.
    \item \textbf{Gradient masking}: Prevents catastrophic forgetting of existing knowledge.
    \item \textbf{Selective unfreezing}: Focuses adaptation capacity on the most relevant parameters.
\end{enumerate}

\section{Conclusion}
\label{sec:conclusion}

We presented AraToken, an Arabic-optimized tokenizer achieving 18\% lower fertility than unnormalized baselines through SentencePiece Unigram training with comprehensive Arabic normalization. We further introduced the Language Extension Pipeline (LEP) for efficiently integrating the tokenizer into Qwen3, reducing evaluation loss from 8.28 to 2.43 within 800 training steps.

\paragraph{Limitations} Our work has several limitations: (1) Experiments are conducted only on Qwen3-0.6B; scaling to larger models requires validation. (2) We do not evaluate on downstream tasks such as question answering or summarization. (3) The training data is limited to Modern Standard Arabic, excluding dialectal Arabic varieties.

\paragraph{Future Work} Future directions include: (1) Extending LEP to larger models (Qwen3-1.7B, 7B), (2) Evaluating on Arabic benchmarks such as ALUE and ArabicGLUE, (3) Adapting the approach for Arabic dialects, and (4) Exploring adapter-based alternatives to full embedding training.

\section*{Acknowledgments}

We thank the SpbU infrastructure team for providing computational resources.

\bibliographystyle{plain}
\bibliography{references}

\end{document}